\documentclass[11pt,a4paper]{article}
\usepackage[hyperref]{acl2017}
\usepackage{times}
\usepackage{latexsym}
\usepackage{booktabs}
\usepackage{amsmath}
\usepackage{amssymb}
\usepackage[defblank]{paralist}
\usepackage{multirow}
\usepackage{flushend}
\usepackage{adjustbox}
\usepackage{enumitem}

\usepackage{url}

\newcommand{\squishlist}{
 \begin{list}{$\bullet$}
  { \setlength{\itemsep}{0pt}
     \setlength{\parsep}{1pt}
     \setlength{\topsep}{1pt}
     \setlength{\partopsep}{0pt}
     \setlength{\leftmargin}{1.5em}
     \setlength{\labelwidth}{1em}
     \setlength{\labelsep}{0.5em} } }
 \newcommand{\squishend}{\end{list}}

\aclfinalcopy 
\def\aclpaperid{3260} 

\title{Cardinal Virtues: Extracting Relation Cardinalities from Text}

\author{Paramita Mirza$^1$, Simon Razniewski$^2$, Fariz Darari$^2$, Gerhard Weikum$^1$ \\ \ \\
$^1$ Max Planck Institute for Informatics\\
$^2$ Free University of Bozen-Bolzano\\
{\tt \{paramita, weikum\}@mpi-inf.mpg.de}\\ 
{\tt \{razniewski, darari\}@inf.unibz.it}}

\begin{document}
\maketitle
\begin{abstract}

Information extraction (IE) from
text has largely focused on relations between individual entities, such as who has won which award.
However, some facts are never fully mentioned, and no IE method has perfect recall.
Thus, it is beneficial to also tap contents about the cardinalities of these relations, for example, how
many awards 
someone has won.
We introduce this novel problem of extracting cardinalities and discuss specific challenges
that set it apart from standard IE.
We present a distant supervision method using conditional random fields.
A preliminary evaluation 
results in precision between 3\% and 55\%, depending on the difficulty of relations.
\end{abstract}


\section{Introduction}
\label{sec:intro}


%

\paragraph{Motivation}
Information extraction (IE) 
can infer relations
between named entities from text (e.g., \cite{NELL:Mitchell2015,delcorro2013clausie,Mausam2012}), 
yielding for example which awards an athlete 
has won,
or instances of family relations like spouses, children, etc.
These methods can be harnessed for 
summarization, question answering (QA), and more.
For populating knowledge bases (KBs),
the IE output is usually cast into subject-predicate-object (SPO)
triples, 
such as
{\em $\langle$BarackObama, hasChild, Malia$\rangle$}, or 
sometimes $n$-ary tuples such as
{\em $\langle$MichaelPhelps, hasWon, OlympicGold, 200mButterfly, 2016$\rangle$}.

IE has focused on capturing full SPO triples (or $n$-ary facts)
with all arguments bound to entities for relation P.
However, 
news, biographies or discussion forums
often 
contain numeric expressions that reveal cardinalities of relations.
Phrases such as ``her two children'' or ``his 28th medal'' are
valuable cues for 
quantifying
the \emph{hasChild} and
\emph{hasWon} relations. 
This can be harnessed in QA
for cases like ``Who won the most Olympic medals?''

An 
important 
application of relation cardinalities is KB curation.
KBs are notoriously incomplete, contain erroneous triples,
and are limited in keeping up with the pace of real-world changes.
For example, a KB may contain only 10 of the 28 Olympic medals that Phelps has won,
or may incorrectly list 3 children for Obama.
Extracting the cardinalities of relations for given subject entities can address all of
these issues.
%
%


Relation cardinalities are disregarded by virtually all IE methods.
Open IE methods \cite{Mausam2012,delcorro2013clausie} 
capture triples (or quadruples)
such as {\em $\langle$Obama, has, two children$\rangle$}.
However, 
there is no way to interpret the numeric expression in the O slot of this triple.
While IE methods that hinge on pre-specified relations
for KB population (e.g., NELL \cite{NELL:Mitchell2015}) can already capture numeric values for explicitly stated attributes such as 
{\em $\langle$Berlin2016attack, hasNumOfVictims, 32$\rangle$}, they are currently not able to learn them.

This paper addresses the novel task of extracting relation cardinalities.
For a given subject entity $s$ and predicate $p$, we aim to
infer the cardinality $|\{\langle S,P,O \rangle ~|~ S=s, P=p\}|$
directly from text, without having to observe any $O$ entities.
This task poses several challenges:
%
%
\begin{itemize}[topsep=0pt,itemsep=0ex,partopsep=0ex,parsep=0ex,leftmargin=*]
\item {\em IE Training.} 
Most IE methods build on seed-based distant supervision.
However, if the underlying KB is not complete, taking the counts of SPO triples for a given SP pair may result in wrong seeds, which can lead to poor patterns.
%

\item {\em Compositionality.} 
The cardinality of an SP pair for a relation may depend on several cardinality mentions. For example, when observing ``Angelina has \textit{two} sons and \textit{three} daughters'', one could infer the children cardinality by summing.

\item {\em Linguistic Variance.} 
In addition to cardinal numbers, cardinality IE should
also pay attention to number-related terms, e.g., ``Angelina gives birth to \textit{twins}'', or
ordinal information, e.g., ``Angelina's \textit{fourth} child'', which can reveal lower bounds
on relation cardinalities. 
%
%
\end{itemize}






%
\paragraph{Approach and Contribution}
Our method learns patterns of phrases that contain cardinal numbers, relying on the distant supervision approach by counting facts for given SP pairs. 
%
Our technical contributions are as follows: \emph{(i)} we 
provide a statistical analysis of numeric information in Wikipedia articles;
\emph{(ii)} we develop a CRF-based extraction method for relation cardinalities
that achieves precision scores of up to 55\%;
\emph{(iii)} we analyze further challenges in this research and outline possible solutions. 
%
%


\section{Related Work}

\paragraph{Knowledge Bases and Information Extraction}
Automated KB construction is a major effort for quite a while. Some approaches, such as YAGO \cite{suchanek2007yago} or DBpedia \cite{dbpedia}, focus on structured parts of Wikipedia, while other approaches such as OLLIE \cite{Mausam2012}, ClauseIE \cite{delcorro2013clausie} or NELL \cite{NELL:Mitchell2015},
focus on unstructured contents across the whole Web. In the latter, usually the schema is also not predefined, thus 
such approaches are 
called Open IE. Most state-of-the-art systems now rely on distant supervision~\cite{distant-sup-earlier,MintzBSJ09}. 

Despite all efforts, KBs are immensely incomplete. For instance, the average number of children per person in Wikidata \cite{vrandevcic2014wikidata} is just 0.02~\cite{vision-AKBC}.

\paragraph{Numbers and Relation Cardinalities}

Numbers in text are an important source of information. Much work has been done on understanding numbers that express temporal information~\cite{ling2010temporal,strotgen2010heideltime}, and more recently, on numbers that express physical quantities or measures, either mentioned in text~\cite{chaganty-liang:2016:P16-1} or in the context of web tables~\cite{yusra-CIKM-2016,ISWC-2016-numeric-values}.

In contrast, numbers that express relation cardinalities have received little attention so far. State-of-the-art Open-IE systems either hardly extract cardinality information 
or fail to extract cardinalities at all. While NELL, for instance, knows 13 relations about the number of casualties and injuries in disasters, they all contain only seed facts and no learned facts. The only prior work we are aware of is of ~\newcite{mirza2016expanding}, who use manually created patterns to mine children cardinalities from Wikipedia. It is shown that with 30 manually crafted patterns and simple filters it is possible to extract 86,227 children-cardinality-assertions with a precision of 94.3\%.

\section{Relation Cardinalities}

\paragraph{Definition}

We define a mention that expresses relation cardinalities as the following:
\emph{``A cardinal number that states the number of objects that stand in a specific relation with a certain subject.''}

Using this definition,
we analyzed how often relation cardinalities occur in Wikipedia.
Relying on the part-of-speech (PoS) tagger of Stanford CoreNLP~\cite{manning2014stanford}, we extracted numbers--i.e., words tagged as \textsc{cd} (cardinal number)--from 10,000 random Wikipedia articles.
The distribution of their named-entity (NE) tags, according to Stanford NE-tagger, is shown in Table~\ref{tbl:CD-tags}.
While temporal-related numbers are the most frequent, around 40\% are classified only as unspecific \textsc{number}.
By manually checking 100 random \textsc{number}s, we observed that 47 are relation cardinalities,\footnote{Among the others are measures, age, or expressions like ``\textit{one} of the...''.} i.e., approximately 18.86\% of all numbers in Wikipedia are relation cardinalities.

We also analyzed the nouns frequently modified by \textsc{number}s, based on their dependency paths, finding \textit{people}, \textit{games}, \textit{children}, \textit{times}, \textit{members} and \textit{seasons} among the top nouns. 
Coarse topic-grouping of the nouns shows that most \textsc{number}s are about sport (\textit{games}, \textit{goals}), followed by artwork (\textit{seasons}, \textit{books}), politics and organization (\textit{members}, \textit{countries}), and family (\textit{children}).

%


\begin{table}
\centering
\begin{adjustbox}{width=0.42\textwidth}
\begin{tabular}{@{}lr@{}}
NE Tag & Frequency \\ \hline
\textsc{date}, \textsc{time}, \textsc{duration}, \textsc{set} (temporal) & 54.28\% \\
\textsc{number} & 40.13\%   \\
\emph{\ \ \ Relation cardinality} & \emph{18.86\%} \\
\textsc{percent} & 2.92\%    \\
\textsc{money} & 2.25\%    \\
\textsc{person}, \textsc{location}, \textsc{organization} & 0.26\%    \\
\textsc{ordinal} & 0.16\%    \\
\end{tabular}
\end{adjustbox}
\caption{NE-tags of numbers in Wikipedia.} 
\label{tbl:CD-tags}
\end{table}



\section{Relation Cardinality Extraction}
Ideally, we would like to make sense of all cardinality statements found in text. However, this would require us to resolve the meaning of a large set of vague predicates, which is in general a difficult task. We thus turn the problem around: given a well defined relation/predicate $p$, a subject $s$ and a corresponding text about $s$, we now try to estimate the relation cardinality (i.e., the count of $\langle s,p,*\rangle$ triples), based on cardinality assertions found in the text. 
We chose four Wikidata predicates that span various domains, \emph{child} (P40), \emph{spouse} (P26), \emph{has part} (P527) of a series of creative works (restricted to novel, book and film series), and \emph{contains administrative territorial entity} (P150). As the text source for subjects of each predicate, we consider sentences containing numbers taken from their respective English Wikipedia articles. 
\paragraph{Methodology} We approach the problem via sequence labelling, i.e., given a sentence containing at least one number, we aim to determine whether each number in the sentence corresponds to the cardinality of a certain relation.
We build a Conditional Random Field (CRF) based model with CRF++~\cite{kudo2005crf++} for each relation, taking as features the context lemmas (window size of 5) around the observed token $t$, along with bigrams and trigrams containing $t$.

To generate the training data, we rely on distant supervision, annotating \textit{candidate numbers}\footnote{Numbers that are not labelled as \textsc{date}, \textsc{time}, \textsc{duration}, \textsc{set}, \textsc{money} and \textsc{percent} by Stanford NE-tagger.} in the text as correct cardinalities whenever they correspond to the exact triple count (count $>0$) found in the knowledge base. 
Otherwise, they are labelled as O (for Others), like the rest of non-number tokens. Table~\ref{tbl:results-vanilla} contains for each considered relation ($p$), the number of subjects (\#$s$) in Wikidata, which have links to English Wikipedia pages and have at least one $\langle s,p,*\rangle$ triple. 

We predict the relation cardinality of a given $\langle s,p\rangle$ pair by selecting the number positively annotated with marginal probability--resulting from forward-backward inference--higher than 0.1, and choosing the one with the highest probability if there are several.

\begin{table*}
\centering
\begin{adjustbox}{width=1.0\textwidth}
\begin{tabular}{lr||c|ccc|ccc}
 & & \textit{baseline} & \multicolumn{3}{c}{\textit{vanilla}} & \multicolumn{3}{|c}{\textit{only-nummod}} \\
$p$ & \#$s$ & P & P & R & F1 & P & R & F1 \\ 
\hline
has part (creative work series) & 261 & .050 & .333 & .316 & .324 & .353 & .316 & .333 \\
contains admin. terr. entity & 18,000 & .034 & .390 & .188 & .254 & .548 & .200 & .293 \\
spouse & 45,917 & 0 & .014 & .011 & .013 & .028 & .017 & .021 \\
child & 35,057 & .112 & .151 & .129 & .139 & .320 & .219 & .260 \\
\hline
child (manual ground truth) & 6,408 & & 0.374 & 0.309 & 0.338 & 0.452 & 0.315 & 0.371 \\
\end{tabular}
\end{adjustbox}
\caption{Number of Wikidata entities as subjects (\#$s$) of each predicate ($p$), and evaluation results on manually annotated randomly selected subjects that have at least an object. 
}
\label{tbl:results-vanilla}
\end{table*}

\paragraph{Experiments} 
Two experimental settings are considered: \textit{vanilla} refers to the distant supervision approach explained above, while for \textit{only-nummod}, we only annotate a candidate number as correct cardinality if it modifies a noun, i.e., there is an incoming dependency relation of label \textit{nummod} according to the Stanford Dependency Parser. This is to exclude numbers as in ``\textit{one} of the reasons...'' from training examples. We also considered a naive baseline, which chooses a random number from a pool of numbers existing in each text about a certain subject. 

Furthermore, to estimate how well KB counts are suited as ground truth, we compare them on the the child relation with the manually-created \textit{number of children} (P1971) property from Wikidata.

\paragraph{Evaluation Results} We manually annotated the evaluation data with the true relation counts, since the knowledge base is highly incomplete, and thus, the triple counts are often incorrect. Whenever the cardinality matches the true count, we also manually inspected how relevant the textual evidence--the context surrounding the cardinal number--is for the observed relation.
Table~\ref{tbl:results-vanilla} shows the performance of our CRF-based method in finding the correct relation cardinality, evaluated on manually annotated 20 (\textit{has part}), 100 (\textit{admin.\ terr.\ entity}) and 200 (\textit{child} and \textit{spouse}) randomly selected subjects that have at least one object.

The random-number baseline achieves a precision of 5\% (\emph{has part}), 3.5\% (\emph{admin.\ territ.\ entity}), 0\% (\emph{spouse}) and 11.2\% (\emph{child}).
Compared to that, especially using \emph{only-nummod}, our method gives encouraging results for \emph{has part, admin.\ territ.\ entity and child}, with 30-50\% precision and around 30\% F1-score. For \emph{spouse}, the performance is significantly lower, reasons are discussed below.
%
%
Furthermore, we can observe that using manual ground truth as training data for the \textit{child} relation can boost performance considerably.
Still, the performance is significantly below the state-of-the-art in fact extraction, where \textit{child} triples can be extracted from Wikipedia text with 96\% precision~\cite{DBLP:conf/icwsm/PalomaresAKR16}. 

\section{Analysis}

A qualitative analysis of the training data and evaluation results revealed three aspects that make extracting relation cardinalities difficult.

\paragraph{Quality of Training Data}
Unlike training data for normal fact extraction, which is generally highly correct (e.g.,\ YAGO claims 95\% precision~\cite{suchanek2007yago}), taking triple counts found in knowledge bases as ground truth generally gives wrong results. For example, our manual evaluation of \textit{child} shows that the triple count from Wikidata is 46\% lower than what the texts assert.

As shown by the last row of Table~\ref{tbl:results-vanilla}, higher quality of training data can considerably boost the performance of cardinality extraction. Unfortunately, manually curated data is generally difficult to obtain. We see two avenues to tackle training data quality:
\begin{enumerate}[topsep=0pt,itemsep=0ex,partopsep=0ex,parsep=0ex,leftmargin=*]
\item \emph{Filtering ground truth.} Instead of taking the counts of all entities as ground truth, one might trade size for quality, e.g., using popular entities only,
as for these there are chances that KBs are more complete.
\item \emph{Incompleteness-resilient distant supervision.} Triple counts in KBs are often lower than what is correct, but rarely too high. Thus, an avenue might be to label all numbers equal or higher than the KB count as correct, instead of only considering the equal ones. Given that different cardinalities could then be labelled as correct, this would require a postprocessing step in which conflicting counts are consolidated.
\end{enumerate}


\vspace{-0.4em}
\paragraph{Compositionality}
Around 16\% of false positives in extracting \textit{child} cardinalities can be attributed to failures in identifying the correct count for, e.g., 
"They have \textit{two} sons and \textit{one} daughter together; he has \textit{four} children from an earlier relationship.'' 
This was also observed for other relations, e.g., ``The Qidong county has \textit{4} subdistricts, \textit{17} towns and \textit{3} townships under its juridiction.''
%
We see two avenues to tackle this problem:

\begin{enumerate}[topsep=0pt,itemsep=0ex,partopsep=0ex,parsep=0ex,leftmargin=*]
\item \emph{Aggregating numbers.} In training data generation, one could label a sequence of number as correct cardinalities if the sum of the numbers is equal to the relation count. In the prediction step, one might sum up all consecutive cardinalities that are labelled with sufficient confidence.
\item \emph{Learning composition rules.} One may try to learn the composition of counts, for instance, that children are composed of sons and daughters, then try to extract the composing cardinalities.
\end{enumerate}

\vspace{-0.4em}
\paragraph{Linguistic Variance}

We observe that for the \textit{spouse} relation, expressing the count with cardinal numbers (``He has married \textit{four} times'') is only found for 4\% of subjects. It is more common to express the count with ordinal numbers, e.g., ``John's \textit{first} wife, Mary, ...'', which allows us to conclude that the spouse-count for John is at least--and most probably more than--one. 
An approach to such relations might be to identify ordinals numbers that express lower bounds of relations. Subsequently, one could reason over these bounds and try to infer relation counts.

Our initial motivation was to make sense of the so far ignored large fraction of numbers that express relation cardinalities. However, we noticed quickly that relation cardinalities are frequently also expressed without numbers at all. This is especially true for the case of count zero, which is mostly expressed using negation (``He never married''), and the count one, which is expressed using indefinite articles (``They have a child'') or the signal-word \textit{only} (``Their only child, James''). Terms such as \textit{twins} or \textit{trilogy} are also ways to express domain-specific relation cardinalities.
We see two avenues to approach this variance:
\begin{enumerate}[topsep=0pt,itemsep=0ex,partopsep=0ex,parsep=0ex,leftmargin=*]
\item \emph{Translation to numbers.} For the 0's and 1's, a possible approach is to translate certain kinds of negation and indefinite articles into explicit numbers (e.g., ``do \textit{not} have \textit{any} children'' $\rightarrow$ ``have 0 children'').
\item \emph{Word similarity with cardinals.} If a word bears high similarity with cardinal numbers, possibly also in other languages such as Latin or Greek, one might consider it as a candidate number.
\end{enumerate}

\section{Conclusion}

In this paper we have introduced the problem of relation cardinality extraction. We believe that relation cardinalities can be useful in a variety of tasks. Our next goal is to make distant supervision incompleteness-resilient and to deal with compositionality, hoping that these can improve the precision of our approach. We also aim to take ordinals into account and to experiment with linguistic transformation for the cases of cardinalities 0 and 1, hoping that these could boost the recall.

A limitation of our work is also that we only focus on Wikipedia articles, assume that all statements are about the article's subject, and just take the statement with the highest confidence. In future work we aim to include a larger article base in combination with named entity recognition, coreference resolution and a truth consolidation step.


\subsection*{Acknowledgments}

We thank Werner Nutt and Sebastian Rudolph for their feedback on an earlier version of this work. We thank the anonymous reviewers for their helpful comments.
This work has been partially supported by the project ``The Call for Recall'', funded by the Free University of Bozen-Bolzano.

\bibliographystyle{acl_natbib}
\bibliography{compl}

\end{document}